\documentclass{article}
\usepackage{spconf,amsmath,graphicx}
\usepackage{algorithm}
\usepackage{algpseudocode}
\usepackage{multirow}
\usepackage{stfloats}
\usepackage{booktabs}
\usepackage{siunitx}
\usepackage{tabularx}
\usepackage[utf8]{inputenc}
\usepackage{subcaption}
\usepackage[hidelinks]{hyperref}

\title{INCPrompt: Task-Aware incremental Prompting for Rehearsal-Free Class-incremental Learning}
\name{Zhiyuan Wang$^{1,2\dag\ddag}$, Xiaoyang Qu$^{2\dag}$, Jing Xiao$^{2}$, Bokui Chen$^{1,3*}$, Jianzong Wang$^{2}$
\thanks{$\dag$ Equal contribution}
\thanks{$\ddag$ Work done as an intern at Ping An Technology (Shenzhen) Co., Ltd }
\thanks{$*$ Corresponding author}
}
\address{$^{1}$Tsinghua Shenzhen International Graduate School, Tsinghua University, China \\
$^{2}$Ping An Technology (Shenzhen) Co., Ltd., Shenzhen, China\\
$^{3}$Peng Cheng Laboratory, Shenzhen, China\\
}

\begin{document}
\maketitle
\begin{abstract}
This paper introduces INCPrompt, an innovative continual learning solution that effectively addresses catastrophic forgetting. INCPrompt’s key innovation lies in its use of adaptive key-learner and task-aware prompts that capture task-relevant information. This unique combination encapsulates general knowledge across tasks and encodes task-specific knowledge. Our comprehensive evaluation across multiple continual learning benchmarks demonstrates INCPrompt’s superiority over existing algorithms, showing its effectiveness in mitigating catastrophic forgetting while maintaining high performance. These results highlight the significant impact of task-aware incremental prompting on continual learning performance.\href{https://github.com/XiaoAI1989/INCPrompt}{https://github.com/XiaoAI1989/INCPrompt}.
\end{abstract}
\section{Introduction}
\label{sec:intro}

Continual learning, the process of acquiring new tasks without erasing knowledge of previously learned ones, presents a substantial hurdle in deep learning. Catastrophic forgetting, the tendency of models to lose knowledge from previous tasks when learning new ones, is a pressing issue that hinders the practical application of sizeable pre-trained vision models \cite{hadsell2020embracing,parisi2019continual,french1999catastrophic}.

In response to these challenges, researchers have proposed various continual learning strategies broadly categorized into two methods: regularization-based and rehearsal-based. Regularization-based methods constrain parameter updates during new task learning to protect old knowledge, typically adding regularization terms to the loss function \cite{li2017learning,kirkpatrick2017overcoming,liu2023fedet}. Rehearsal-based methods maintain some previous task data, mixing them with new task data during training \cite{lopez2017gradient,rebuffi2017icarl,wang2023shoggoth}.

\begin{figure}[htb]

    \centering
    \begin{subfigure}[b]{0.45\textwidth}
        \includegraphics[width=\textwidth,keepaspectratio]{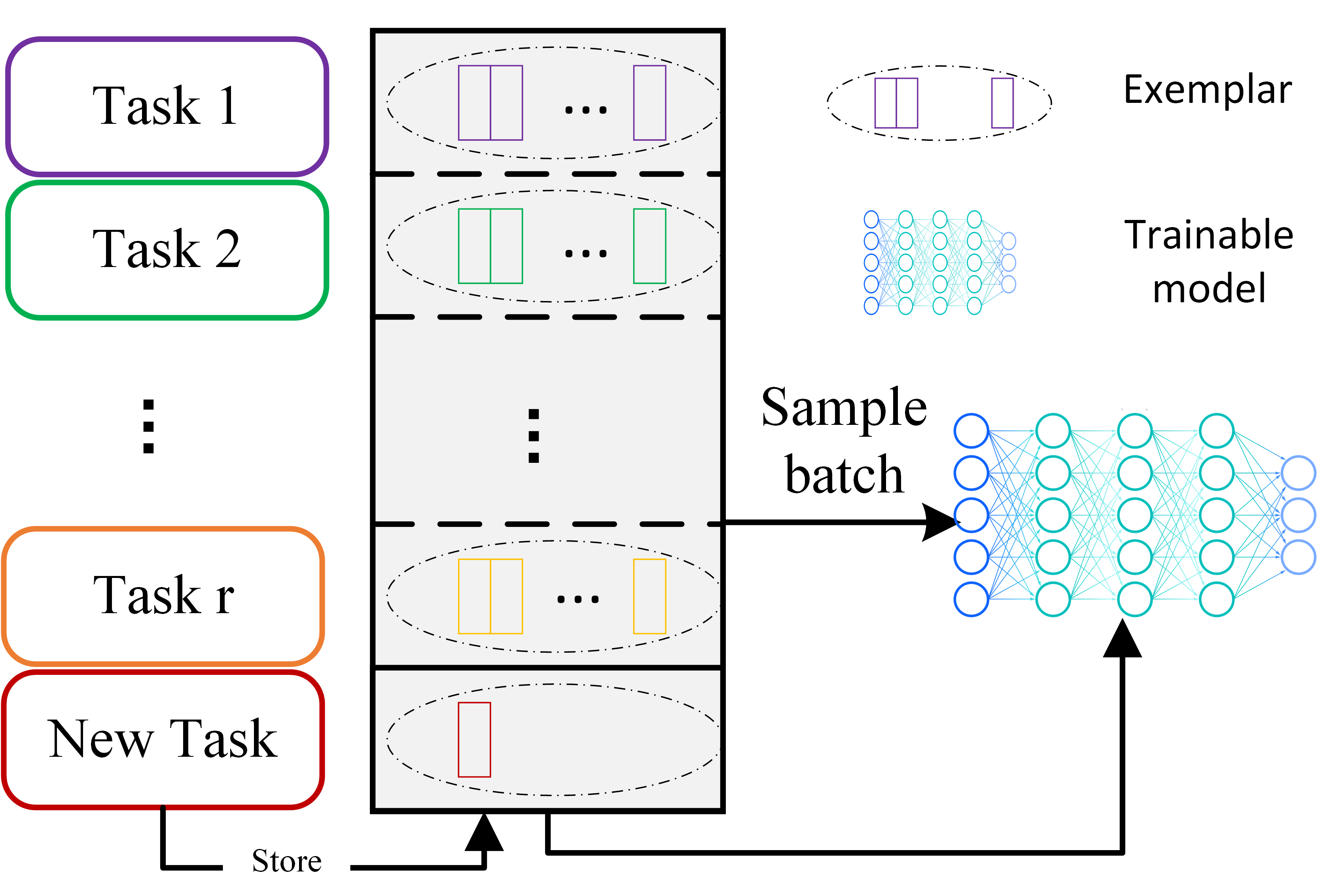}
        \caption{Traditional rehearsal-based methods can be computationally expensive and inefficient due to storing and accessing large amounts of data.}
        \label{fig:why_a}
    \end{subfigure}
    \hfill
    \begin{subfigure}[b]{0.45\textwidth}
        \includegraphics[width=\textwidth,keepaspectratio]{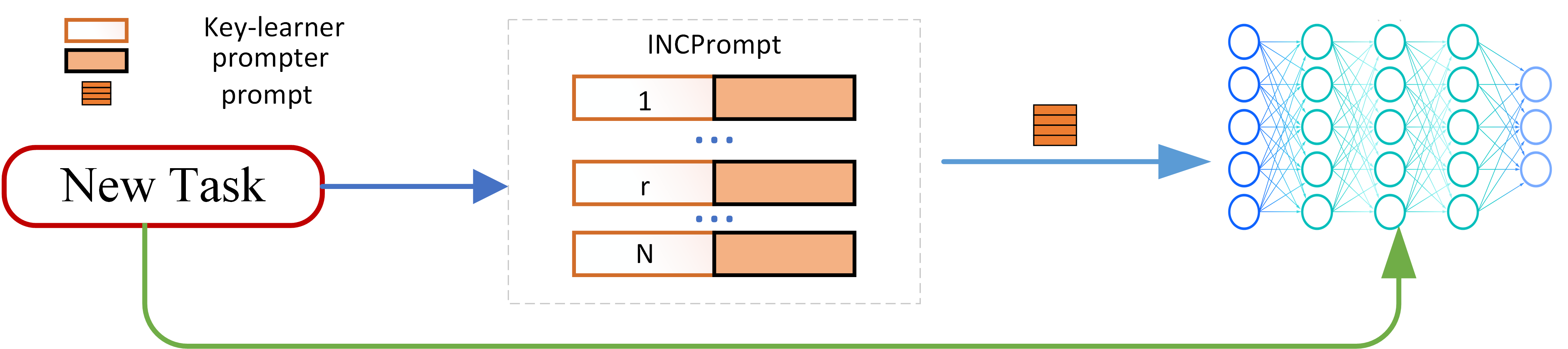}
        \caption{INCPrompt offers a more efficient strategy. It utilizes a prompt-generating model. This model stores task-specific knowledge, removing the need for a rehearsal buffer.}
        \label{fig:why_b}
    \end{subfigure}
    \caption{Comparison of traditional methods and INCPrompt.}
    \label{fig:why}
    \vspace{-15pt}
\end{figure}

Prompt learning strategically modifies the input text, providing language models with insights specific to the task. The design of effective hand-crafted prompts can be complex, often requiring heuristic approaches. Soft prompting \cite{lester2021power} and prefix tuning \cite{li2021prefix} circumvent this by introducing learnable parameters that guide the model's behavior, simplifying the adaptation of large language models to specific tasks. There have been substantial advancements in this field, such as the L2P \cite{wang2022learning} method, which learns a mapping from task descriptions to prompts. The DualPrompt \cite{wang2022dualprompt} approach utilizes dual prompts to enhance model control. However, these methods lack our proposed method's dynamic, task-specific adaptability.

To address these challenges, we advance the field through a series of novel contributions, detailed below:
\begin{itemize}
\item We introduce INCPrompt, which balances new task adaptability and knowledge retention from previous tasks. 
\item It incorporates a regularization term to maintain a balance between adaptation to reduce overfitting and encourage feature extraction.
\item We conduct comprehensive experimental evaluations against other methods across various benchmarks.
\end{itemize}

\section{The Proposed Method}
\label{sec:system}

\subsection{Problem Setting }
\label{ssec:prob}

\begin{figure*}[htb]
    \centering
    \begin{subfigure}[b]{0.6\textwidth}
        \includegraphics[width=\textwidth,keepaspectratio]{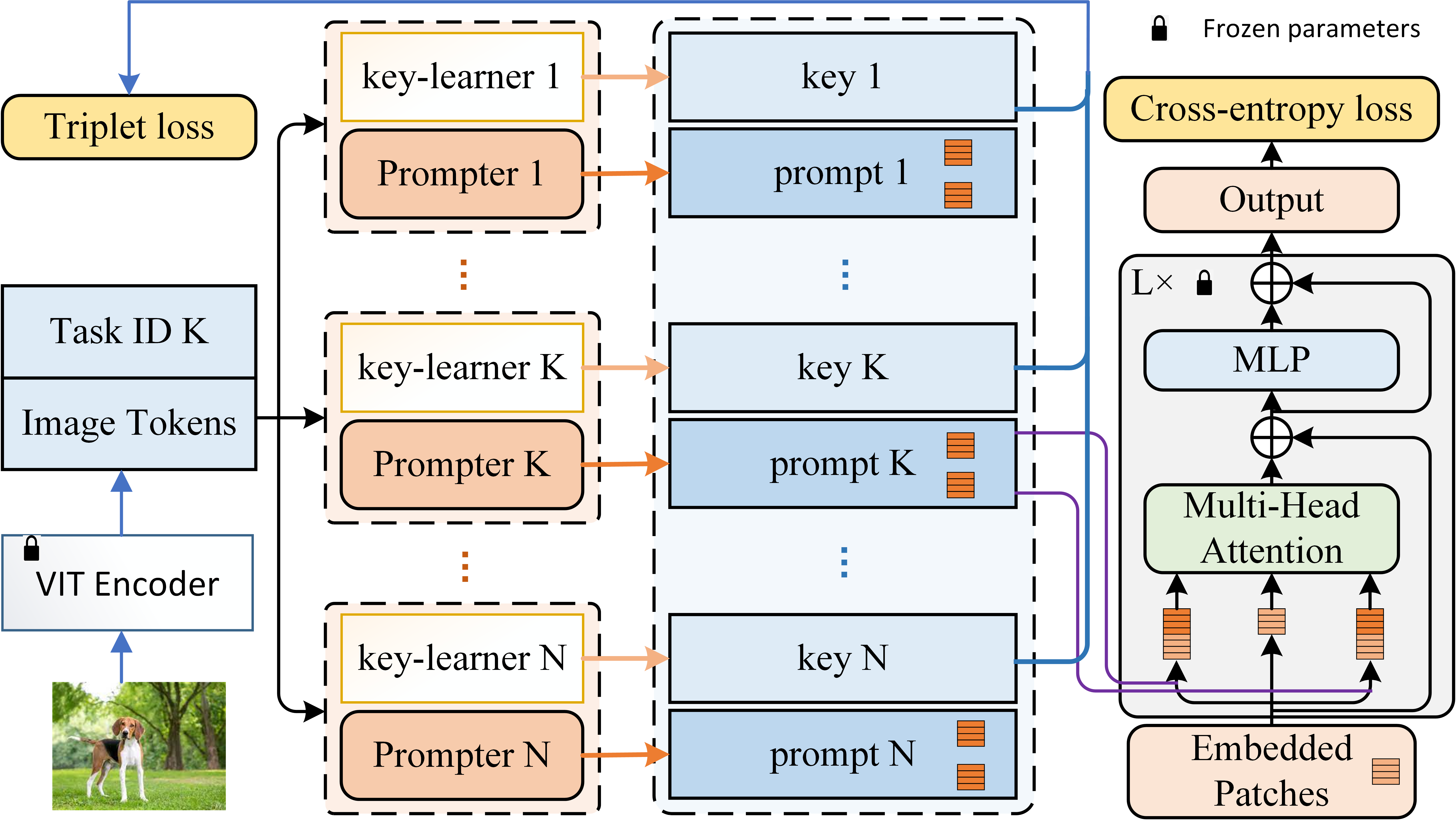}
        \caption{Training procedure}
        \label{fig:how_a}
    \end{subfigure}
    \hfill
    \begin{subfigure}[b]{0.38\textwidth}
        \includegraphics[width=\textwidth,keepaspectratio]{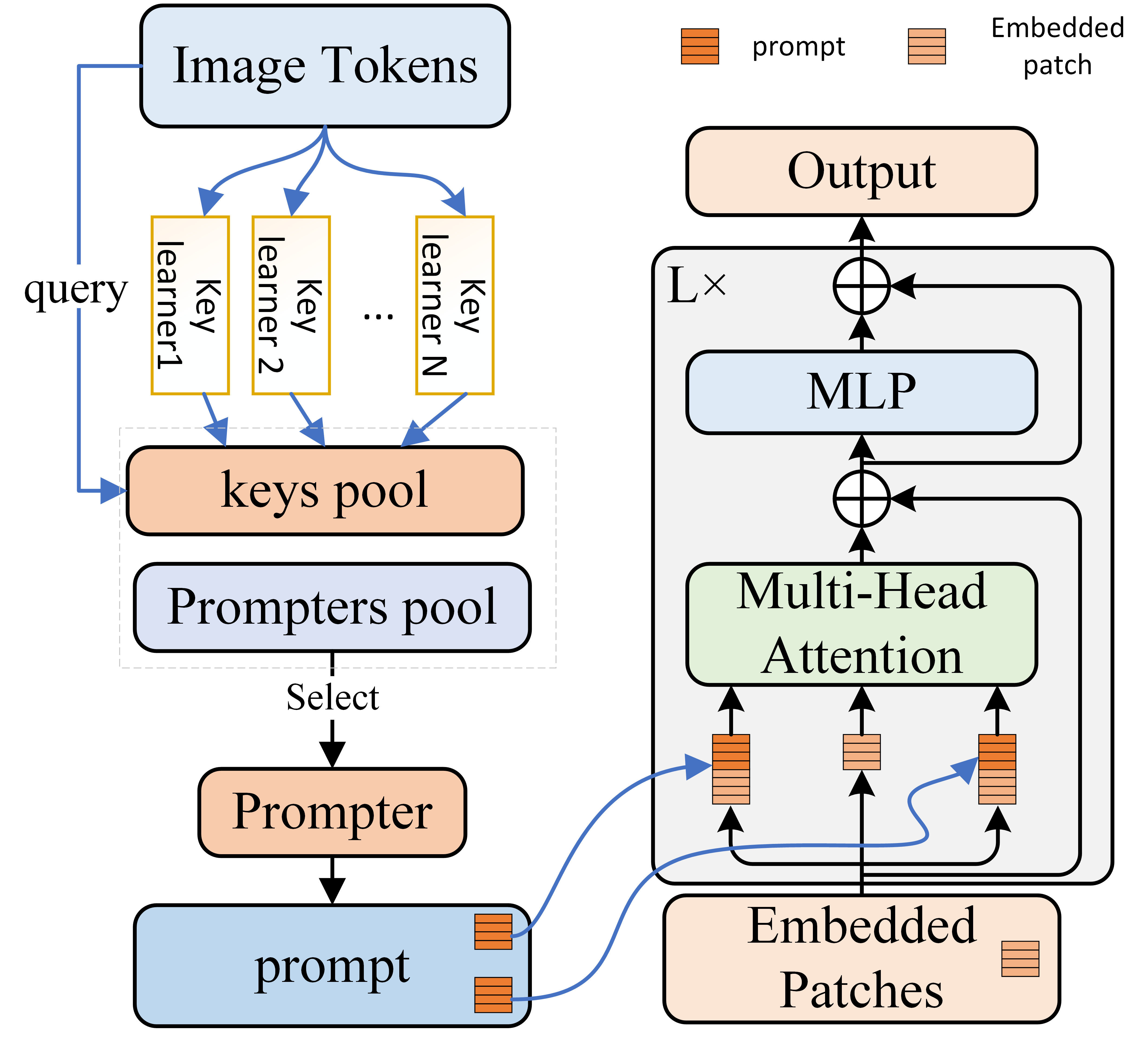}
        \caption{Inference procedure}
        \label{fig:how_b}
    \end{subfigure}
    \caption{The architecture of our algorithms.}
    \label{fig:how}
    \vspace{-15pt}
\end{figure*}

\vspace{-0.5em}

A model is exposed to a sequence of $N$ tasks $D = {D_1, ..., D_N}$ in our setting. Each task exclusively contains different semantic object categories, with the objective being to progressively acquire the ability to categorize newly introduced object classes while preserving the classification accuracy of object classes learned in the past. Here's a formalized definition of the continual learning problem:

For each task $D_t$ consisting of pairs $(x_{i,t}, y_{i,t})$ with $i$ ranging from 1 to $n_t$, a model $f_{\theta}$, parameterized by $\theta$, aims to infer the label $y_{i,t} \in Y$ for any input $x_{i,t} \in X$, thereby learning the function from $X$ to $Y$. 
We define the task-specific loss $L_{\text{task}}$ as the cross-entropy loss between the predicted labels and the true labels:
\begin{equation}
L_{\text{task}} = -\sum_{i=1}^{n_t} \sum_{c=1}^{|Y|} y_{i,t,c} \log(f_{\theta}(x_{i,t})_c)
\end{equation}
where $y_{i,t,c}$ is an indicator variable, 1 if the class label $c$ is the correct classification for the observation $x_{i,t}$, and 0 otherwise. It should be emphasized that the data from earlier tasks is inaccessible when training subsequent tasks.

\subsection{Key Learner}
\label{ssec:mp}

The Key-Learner implements an attention-based transformation to the input tokens. The transformed input is then processed by a self-attention mechanism, which is mathematically represented as:

\begin{equation}
\text{Attn}(Q_i, K_i, V_i) = \text{softmax}\left(\frac{Q_iK_i^T}{\sqrt{d_{k_i}}}\right)V_i
\end{equation}
In this updated equation, $Q_i$, $K_i$, and $V_i$ represent the query, key, and value of the output tokens from the Vision Transformer encoder, and $d_{k_i}$ is the dimensionality of the queries and keys. The subscript $i$ denotes that these are specific output tokens from the ViT.

This triplet loss function maximizes the softmax probability of a vector's dot product with a positive sample compared to negative samples. It's defined as:

\begin{equation}
L_{\text{Triplet}} = \sum_{i=1}^{N} [ | T_k^i - K_i^a |_2^2 - | T_k^i - K_i^n|_2^2 + \alpha ]+
\end{equation}
In this equation, $T_k^i$ represents the output tokens for a given input. $K_i^a$ represents the key output from the current key-learner as an anchor point. $K_i^n$ represents the key output from a different key-learner with the highest similarity to the output tokens $T_k^i$ among all other key-learners. $| \cdot |_2$ denotes the Euclidean distance, $[ \cdot ]+$ denotes the operation of taking the maximum between the argument and 0, and $\alpha$ is a margin parameter that is used to separate positive and negative pairs, which can be tuned based on the specific task.

This technique prevents overfitting and encourages feature extraction with a penalty to the $L_1$ norm of the key. It's defined as:

\begin{equation}
L_{\text{reg}} = \sum_{i=1}^{N} ||K_i^a||_1 
\end{equation}
In this equation, $||K_{i}^a||_1$ denotes the L1 norm of the key output from the current key-learner. A higher regularization loss will be obtained if the two have a lower difference. This term encourages the model to keep the $K_P$ different from output tokens and thus helps to prevent overfitting. The total loss of key-learner is denoted as:
\begin{equation}
L_{key} =  \lambda_{\text{reg}} \cdot L_{reg}+L_{Triplet}
\end{equation}
$\lambda_{\text{reg}}$ is a hyper-parameter
and serves as a regularization coefficient that balances the L1-norm regularization term $L_{reg}$ and the triplet loss $L_{Triplet}$.

\subsection{Task-aware Prompter}
\label{ssec:tp}

Moreover, we utilize an ensemble of feedforward layers with a non-linear activation function, termed the `prompter,' to dynamically synthesize the key ($P_k$) and value ($P_v$) components of the prompt $P$. The generation of the prompt by the prompter can be represented by the following equation:

\begin{equation}
P = f_{\text{prompter}}(x_{\text{token}})
\end{equation}
Here, $f_{\text{prompter}}$ corresponds to the prompter function, which receives an image token as input. 

In our proposed INCPrompt method, we integrate an external prompt $P$ into the computation process. The prompt $P$ consists of a key component $P_k$ and a value component $P_v$. The modified attention computation becomes:

\begin{equation}
Att(Q,K \oplus P_k,V \oplus P_v) = f(\frac{Q(K \oplus P_k)^T}{\sqrt{d_k}})(V \oplus P_v)
\end{equation}
where $f(.)$ denotes the softmax function, $\oplus$ denotes the concatenation operation.

In this way, the transformer model can consider the original input and the external prompt when computing the attention. This mechanism ensures that the generated prompts are adaptive and can effectively incorporate task-relevant information into the attention mechanism of the transformer model.

\subsection{Algorithm Architecture }
\label{ssec:arch}

\begin{algorithm}[htb]
\caption{Task-Aware Incremental Prompting}
\label{alg:MetaPrompt}
\begin{algorithmic}[1]
\State \textbf{Input:} Tokens from ViT $x_{token}$, Task ID $x_{id}$
\State \textbf{Output:} Final prompter (p), Contrastive loss (L) if in training mode
\State \textbf{Initialize:} Pre-trained Vision Transformer (V), Key-Learners (K), Task-Prompters (T), Image dataset (D),
    \If{Train}
        \State $K_{p}, K_{n} \gets K(x_{id},x_{token})$ 
        \Comment{Generate pos. and neg.keys}
        \State $L \gets L_{key}(K_{p}, K_{n}, t)$ 
        \Comment{Compute key loss}
        \State $P \gets T(x_{id},x_{token})$ 
        \Comment{Get Task-prompter output}
    \EndIf
    \If{Eval}
        \State $K_{All} \gets K(x_{token})$ 
        \Comment{Generate all keys}
        \State $idx \gets argmax(Sim(K_{All}, x_{token}))$ 
        \Comment{Find the best matching K}
        \State $P\gets T(x_{id},x_{token})$ 
        \Comment{Get the output of the Prompter}

    \EndIf
    \State $p \gets Divide(P)$ 
    \Comment{Divide the prompts into key and value}
    \If{Train}
        \State \Return $p, l$
    \Else
        \State \Return $p, 0$
    \EndIf

\end{algorithmic}
\vspace{-0.3em}
\end{algorithm}

Consequently, the loss function of the INCPrompt is computed as the sum of the components mentioned above:

\begin{equation}
L = L_{task} + L_{key}
\end{equation}
By minimizing this loss function, INCPrompt can effectively learn task-specific and retain task-invariant knowledge, improving performance in continual learning tasks, as shown in Algorithm \ref{alg:MetaPrompt}.

\section{evaluation}
\label{sec:eval}
\subsection{Experiment Setup }
\label{ssec:exp}

\begin{figure}[b]
    \begin{subfigure}{.45\linewidth}
        \centering
        \includegraphics[width=\linewidth]{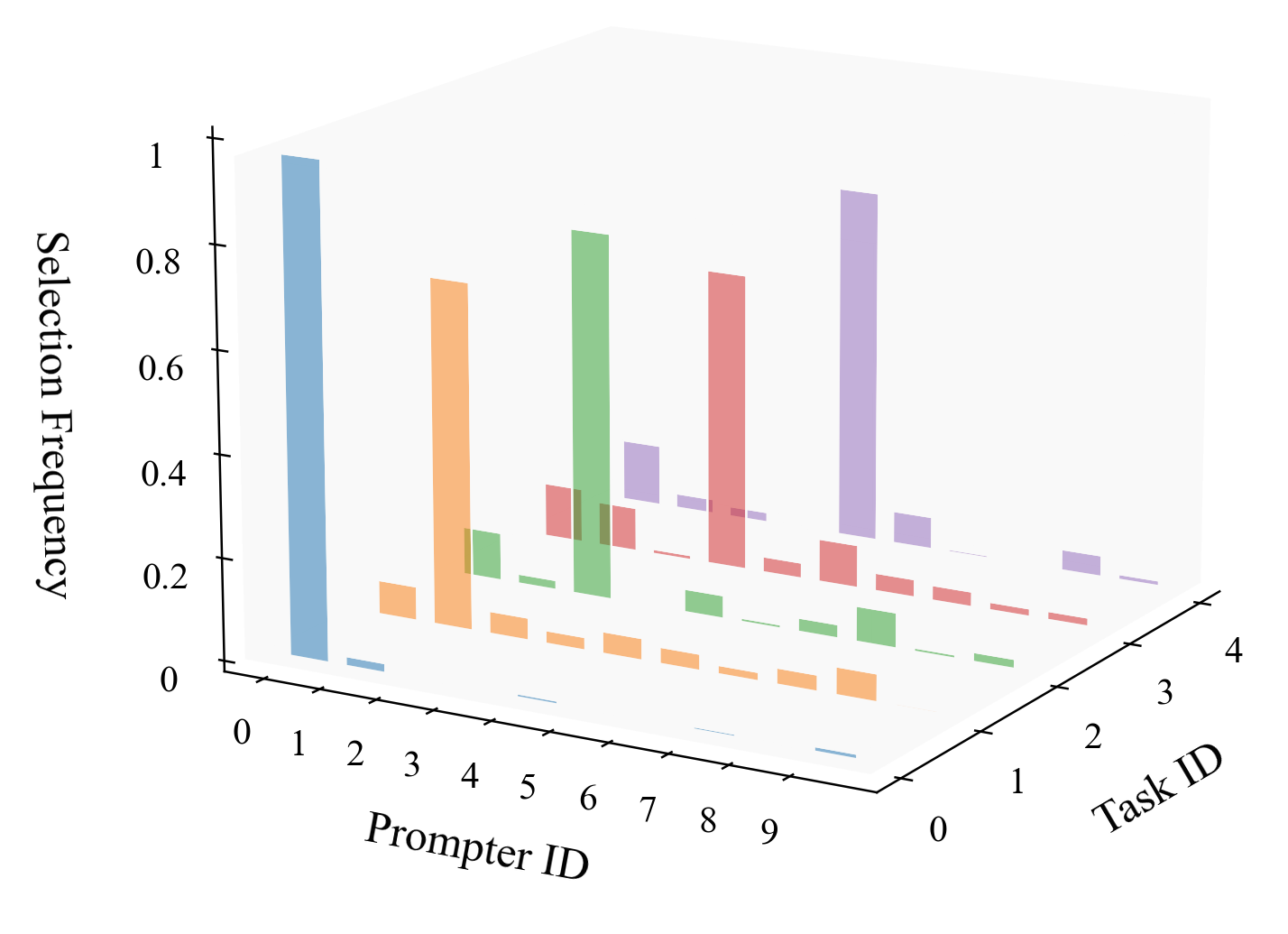}
        \label{fig:sub1}
    \end{subfigure}
    \begin{subfigure}{.45\linewidth}
        \centering
        \includegraphics[width=\linewidth]{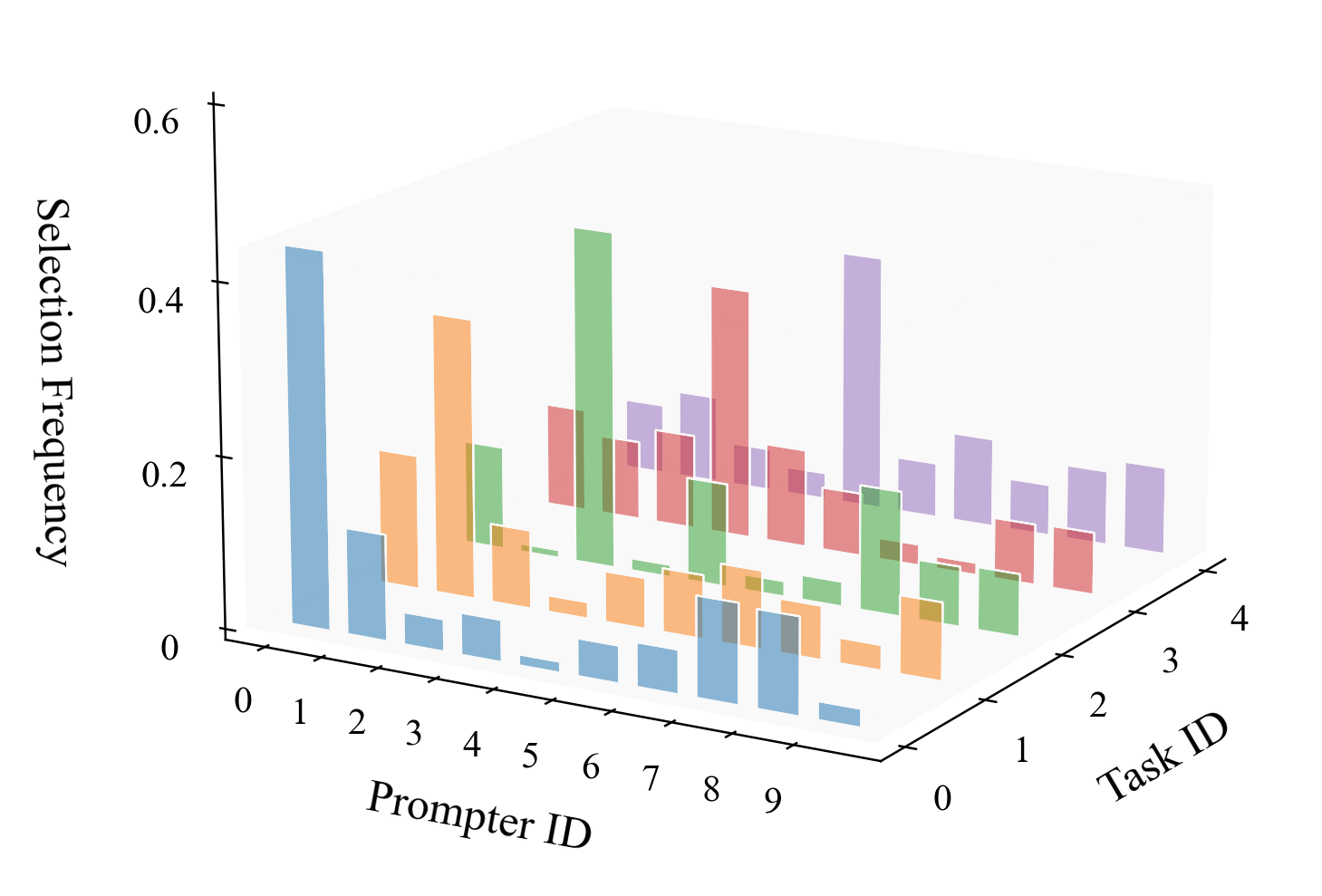}
        \label{fig:sub2}
    \end{subfigure}
    \vspace{-0.3em}
    \caption{Histograms depicting prompter selection for (left) Split CIFAR-100 and (right) ImageNet-R. ImageNet-R has a higher level of intra-task similarity compared to CIFAR100. On the other hand, CIFAR100 benefits more from prompts that are more specific to individual tasks. For better readability, we only display the first five tasks.}
    \label{fig:res}
    \vspace{-0.3em}
\end{figure}

To illustrate our primary findings, we employ the Split CIFAR-100 and Split ImageNet-R benchmark, an adaptation of ImageNet-R \cite{hendrycks2021many}, comprising 200 classes randomly divided into ten tasks, each containing 20 types. It is sufficiently robust to expose a high forgetting rate in advanced CL methods during class-incremental learning in these two datasets.

As shown in Table \ref{tab:comparison}, to ensure a thorough and unbiased experiment, we initially compare our proposed method, INCPrompt, with techniques based on regulation, rehearsal, and prompting approaches. We have chosen characteristic methods like EWC \cite{kirkpatrick2017overcoming}, LwF \cite{li2017learning}, ER \cite{chaudhry2019tiny}, GDumb \cite{prabhu2020gdumb}, BiC \cite{wu2019large}, DER++ \cite{buzzega2020dark}, Co$^{2}$L \cite{cha2021co2l}, L2P \cite{wang2022learning}, and DualPrompt \cite{wang2022dualprompt}, from all categories. To provide a more transparent illustration, we include FT-seq, representing a straightforward sequential training method, and Upper-bound, representing traditional fine-tuning encompassing all tasks. Notably, the accuracy figures reported for our INCPrompt methodology are derived under the specific conditions of employing a prompt length of 20 and a prompt depth of 6. 

\begin{table*}[!htb]
\centering
\caption{Outcomes in continual learning (where task ID remains unavailable during testing). The methods are compared and grouped according to their buffer sizes. A buffer size of zero indicates that no rehearsal was utilized, rendering most state-of-the-art techniques inapplicable.}
\label{tab:comparison}
\begin{tabular}{|c|c|cc|c|cc|}
\hline
\textbf{Method} & \textbf{Dataset Buffer} & \multicolumn{2}{c|}{\textbf{Split CIFAR-100}} & \textbf{Dataset Buffer} & \multicolumn{2}{c|}{\textbf{Split ImageNet-R}} \\
& & \textbf{Avg. Acc (↑)} & \textbf{Forgetting (↓)} & & \textbf{Avg. Acc (↑)} & \textbf{Forgetting (↓)} \\
\hline
ER \cite{chaudhry2019tiny} & 1000 & 67.87±0.57 & 33.33±1.28 & 1000 & 55.13±1.29 & 35.38±0.52 \\
BiC \cite{wu2019large} & 1000 & 66.11±1.76 & 35.24±1.64 & 1000 & 52.14±1.08 & 36.70±1.05 \\
GDumb \cite{prabhu2020gdumb} & 1000 & 67.14±0.37 & - & 1000 & 38.32±0.55 & - \\
DER++ \cite{buzzega2020dark} & 1000 & 61.06±0.87 & 39.87±0.99 & 1000 & 55.47±1.31 & 34.64±1.50 \\
Co$^{2}$L \cite{cha2021co2l} & 1000 & 72.15±1.32 & 28.55±1.56 & 1000 & 53.45±1.55 & 37.30±1.81 \\
\hline
ER \cite{chaudhry2019tiny} & 5000 & 82.53±0.17 & 16.46±0.25 & 5000 & 65.18±0.40 & 23.31±0.89 \\
BiC \cite{wu2019large} & 5000 & 81.42±0.85 & 17.31±1.02 & 5000 & 64.63±1.27 & 22.25±1.73 \\
GDumb \cite{prabhu2020gdumb} & 5000 & 81.67±0.02 & - & 5000 & 65.90±0.28 & - \\
DER++ \cite{buzzega2020dark} & 5000 & 83.94±0.34 & 14.55±0.73 & 5000 & 66.73±0.87 & 20.67±1.24 \\
Co$^{2}$L \cite{cha2021co2l} & 5000 & 82.49±0.89 & 17.48±1.80 & 5000 & 65.90±0.14 & 23.36±0.71 \\
\hline
FT-seq & 0 & 33.61±0.85 & 86.87±0.20 & 0 & 28.87±1.36 & 63.80±1.50 \\
EWC \cite{kirkpatrick2017overcoming} & 0 & 47.01±0.29 & 33.27±1.17 & 0 & 35.00±0.43 & 56.16±0.88 \\
LwF \cite{li2017learning} & 0 & 60.69±0.63 & 27.77±2.17 & 0 & 38.54±1.23 & 52.37±0.64 \\
L2P \cite{wang2022learning} & 0 & 83.86±0.28 & 7.35±0.38 & 0 & 61.57±0.66 & 9.73±0.47 \\
DualPrompt \cite{wang2022dualprompt} & 0 & 84.01±0.28 & 7.01±0.21 & 0 & 71.13±0.42 & 6.79±0.12 \\
\textbf{Ours} & 0 & 85.03±0.21 & 6.64±0.18 & 0 & 73.22±0.45 &  6.47±0.16 \\
\hline
Upper-bound & - & 90.85±0.12 & - & - & 79.13±0.18 & - \\
\hline
\end{tabular}
\end{table*}

\subsection{Results Analysis}
\label{ssec:analysis}

\begin{figure}[htb]
    \begin{minipage}{.45\linewidth}
        \centering
        \includegraphics[width=\linewidth]{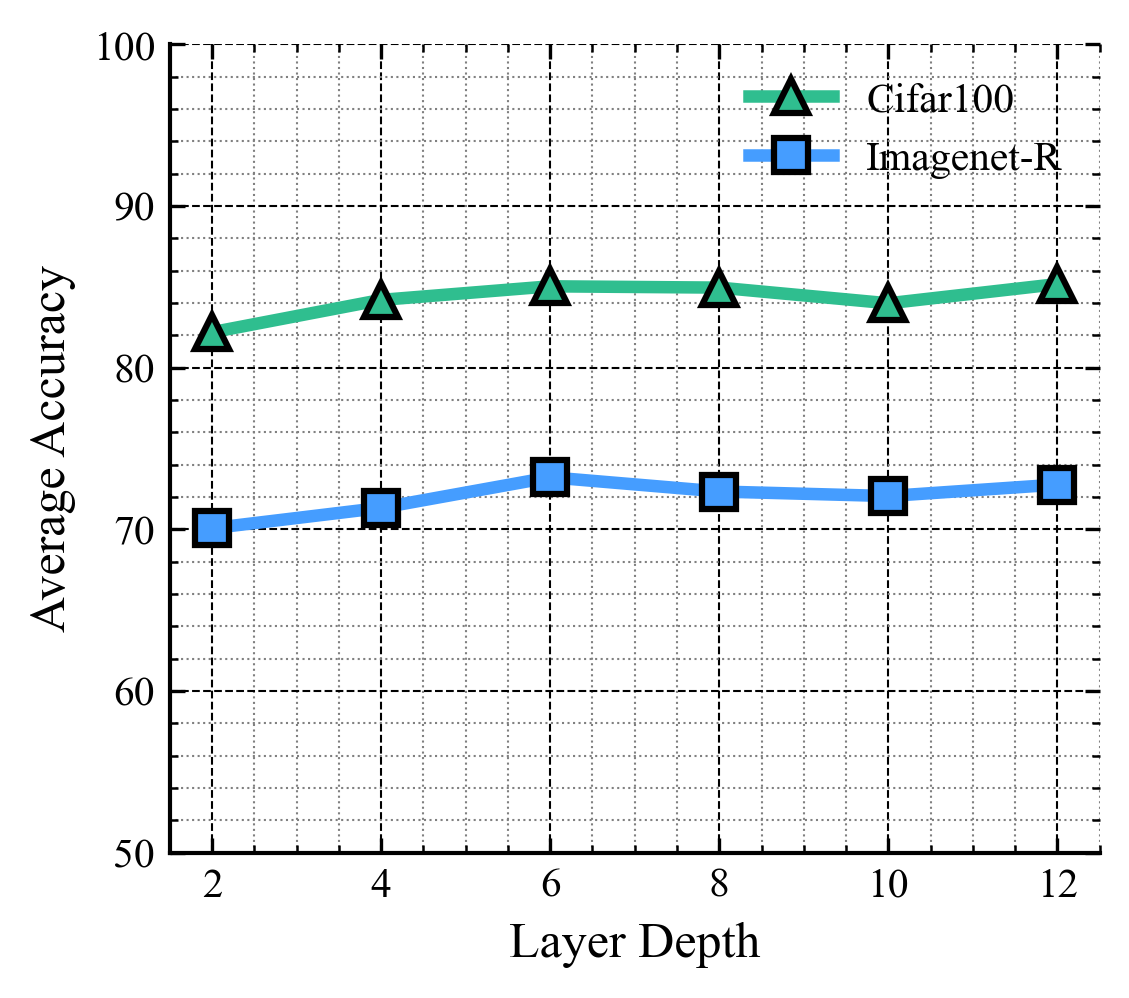}
        \caption{The relationship between the accuracy and the prompt layer depth of ViT blocks. (prompt length=20)}
        \label{fig:img1}
    \end{minipage}
    \hspace{1em} 
    \begin{minipage}{.45\linewidth}
        \centering
        \includegraphics[width=\linewidth]{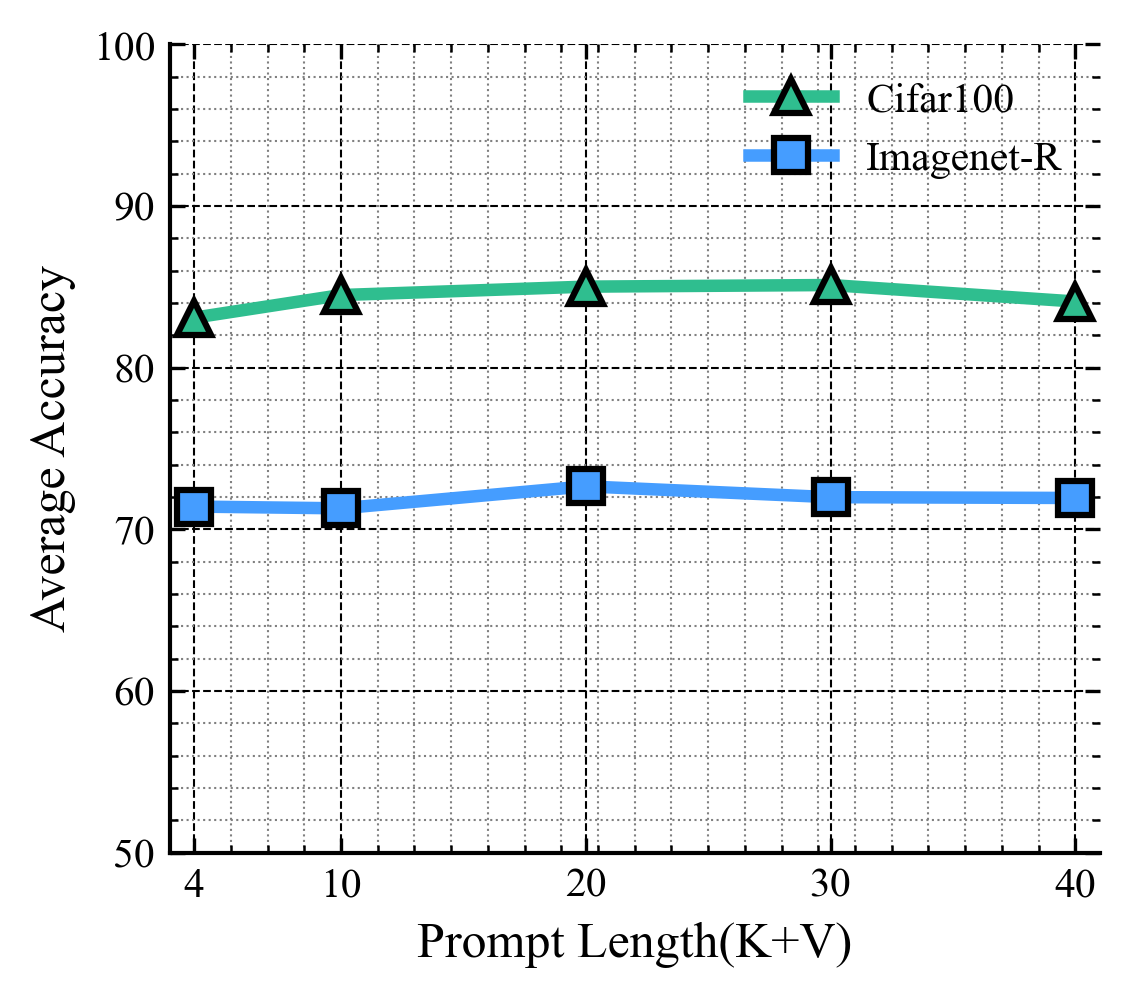}
        \caption{The relationship between the accuracy and the length of the prompt. (layer depth=5)}
        \label{fig:img2}
    \end{minipage}

\end{figure}

Table \ref{tab:comparison} summarizes the methods compared on the two datasets. Our proposed approach, INCPrompt, consistently outperforms all competing methods, including those that do not involve rehearsal and those that rely on rehearsal. Notably, when the buffer size is set to 5000, all rehearsal-based methods exhibit performance similar to GDumb's. This observation suggests that the implementation of rehearsal-based continual learning may not offer substantial performance improvement over supervised training. Figure \ref{fig:img1} shows that prompts added to the 5th MSA layers yield the most effective results. Figure \ref{fig:img2} aims to highlight that, as prompt length increases, accuracy reaches a saturation point.

\section{Conclusion}

\label{sec:typestyle}
This work introduces INCPrompt, a unique continual learning approach leveraging dynamic, task-specific prompts to counter catastrophic forgetting. This innovative blend of key-learners and task-aware prompters enables INCPrompt to maintain general and task-specific knowledge. Testing on various continual learning datasets confirms its balance between new task adaptability and old task knowledge preservation.

\section{ACKNOWLEDGEMENT}
\label{sec:ack}
\vspace{-0.5em}
Sponsored by the Key Research and Development Program of Guangdong Province under grant No.2021B0101400003, Tsinghua-Toyota Joint Research Fund.

\vfill\pagebreak
\label{sec:refs}

\bibliographystyle{IEEE}
\bibliography{refs}

\end{document}